\documentclass{article}
\usepackage{ijcai18}

\usepackage{times}
\usepackage{xcolor}
\usepackage{soul}
\usepackage[utf8]{inputenc}
\usepackage[small]{caption}

\usepackage{multirow}
\usepackage[american]{babel}
\usepackage{graphicx}
\usepackage{amsmath}
\usepackage{amssymb}
\usepackage{float}
\usepackage{url}

\graphicspath{{figures/}}

\title{When Image Denoising Meets High-Level Vision Tasks:  \\ A Deep Learning Approach}

\author{
Ding Liu$^1$, 
Bihan Wen$^1$, 
Xianming Liu$^2$, 
Zhangyang Wang$^3$, 
Thomas S. Huang$^1$ 
\thanks{ Ding Liu and Thomas S. Huang's research work was supported
by the U.S. Army Research Office under Grant W911NF-15-1-0317.}
\\ 
$^1$ University of Illinois at Urbana-Champaign, USA \\
$^2$ Facebook Inc.
$^3$ Texas A\&M University, USA \\
\{dingliu2,bwen3,t-huang1\}@illinois.edu,
xmliu@fb.com,
atlaswang@tamu.edu
}

\begin{document}

\maketitle

\begin{abstract}
Conventionally, image denoising and high-level vision tasks are handled separately in computer vision.
In this paper, we cope with the two jointly and explore the mutual influence between them. 
First we propose a convolutional neural network for image denoising which achieves the state-of-the-art performance. 
Second we propose a deep neural network solution that cascades two modules for image denoising and various high-level tasks, respectively, and use the joint loss for updating only the denoising network via back-propagation.
We demonstrate that on one hand, the proposed denoiser has the generality to overcome the performance degradation of different high-level vision tasks.
On the other hand, with the guidance of high-level vision information, the denoising network can 
generate more visually appealing results. 
To the best of our knowledge, this is the first work investigating the benefit of exploiting image semantics simultaneously for image denoising and high-level vision tasks via deep learning.
The code is available online\footnote{\url{https://github.com/Ding-Liu/DeepDenoising}}.
\end{abstract}

\section{Introduction}

\begin{figure}[th!]
	\centering	
	\begin{tabular}{c@{\hskip 0.5mm}c}
		\multicolumn{2}{@{\hskip 0.5mm}c}{\includegraphics[width=0.9\linewidth]{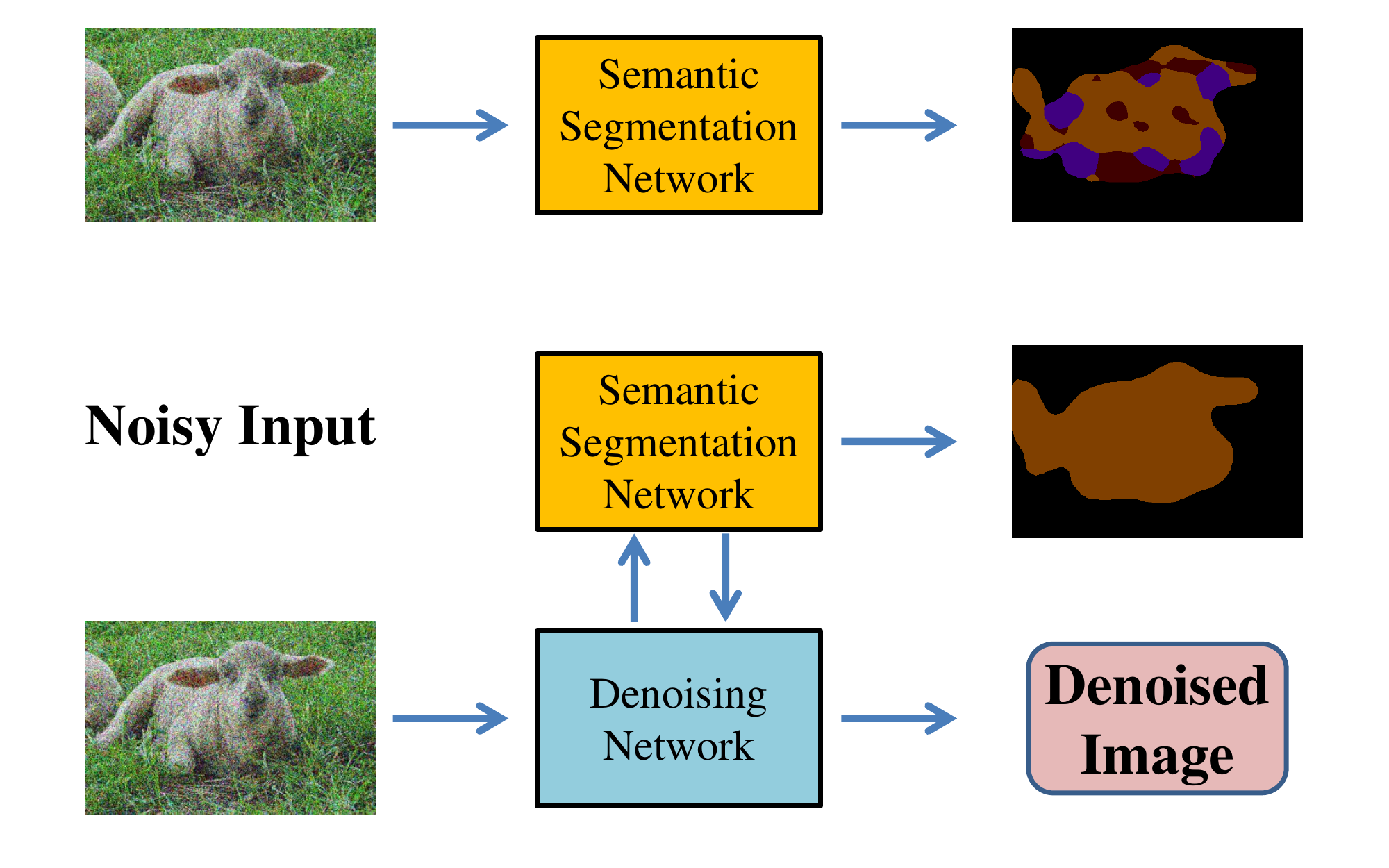}} \\ [-10pt]
		\multicolumn{2}{c}{ (a) } \\
		\includegraphics[width=0.45\linewidth, trim=210 305 180 45,clip]{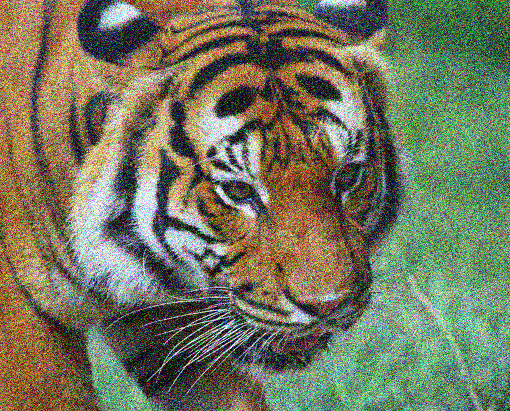} &
		\includegraphics[width=0.45\linewidth, trim=210 305 180 45,clip]{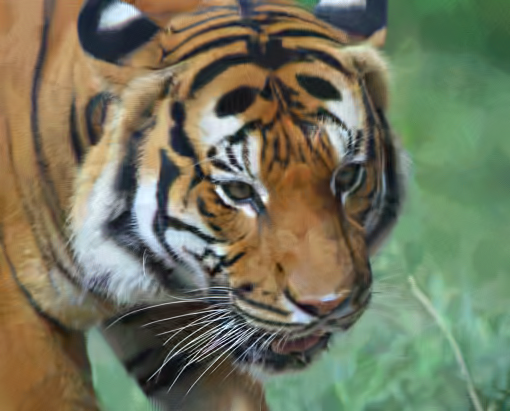}	\\
		Noisy input & CBM3D \\
		\includegraphics[width=0.45\linewidth, trim=210 305 180 45,clip]{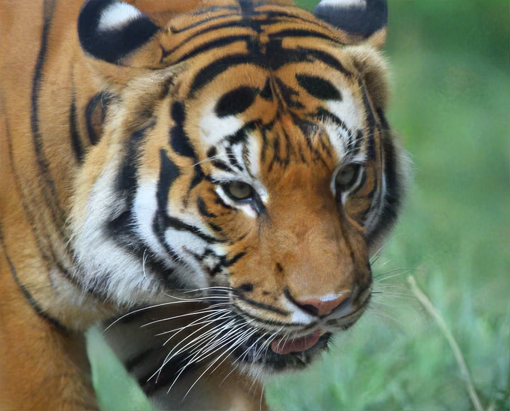} &
		\includegraphics[width=0.45\linewidth, trim=210 305 180 45,clip]{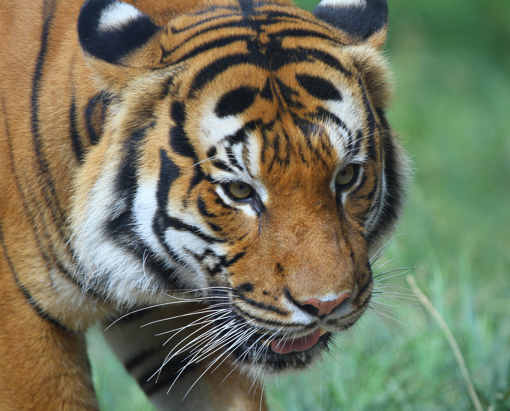}  \\		
		Our method & Ground truth \\
		\multicolumn{2}{c}{ (b) } 			
	\end{tabular}
	\vspace{-5mm}	
	\caption{ (a) Upper: conventional semantic segmentation pipeline; lower: our proposed framework for joint image denoising and semantic segmentation. (b) Zoom-in regions of a noisy  input, its denoised estimates using CBM3D and our proposed method, as well as its ground truth.
	}
	\label{fig:intro}
	\vspace{-5mm}
\end{figure}

\begin{figure*}[th!]
	\centering
	\includegraphics[width=0.85\linewidth]{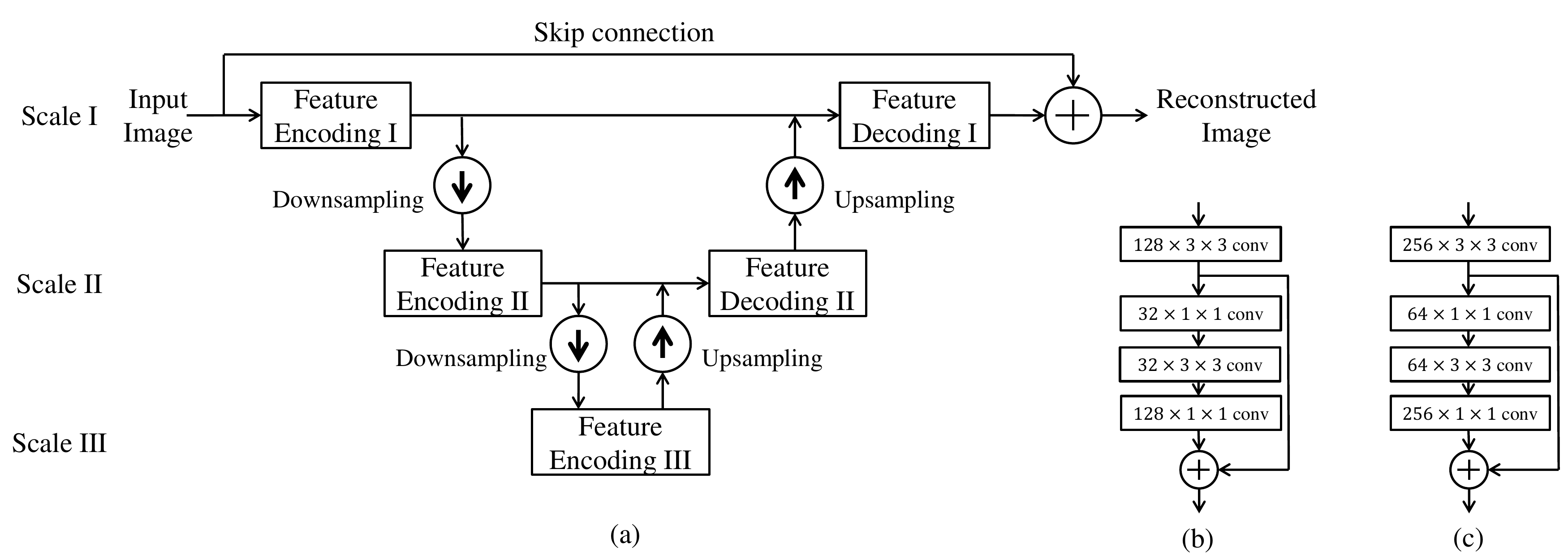}
	\vspace*{-4mm}
	\caption{(a) Overview of our proposed denoising network. (b) Architecture of the feature encoding module. (c) Architecture of the feature decoding module.}
	\label{fig:denoising}
	\vspace*{-5mm}
\end{figure*}

A common approach in computer vision is to separate low-level vision problems, such as image restoration and enhancement, from high-level vision problems, and solve them independently.
In this paper, 
we make their connection by showing
the mutual influence between the two, i.e., visual perception and semantics, and propose a new perspective of solving both the low-level and high-level computer vision problems in a single unified framework, as shown in Fig.~\ref{fig:intro}(a).  

Image denoising, as one representative of low-level vision problems, is dedicated to recovering the underlying image signal from its noisy measurement.
Classical image denoising methods take advantage of local or non-local structures presented in the image \cite{Aharon2006,dabov2007image,Mairal2009,NCSR,WNNM,PGPD}.
More recently, a number of deep learning models have been developed for image denoising which demonstrated superior performance~\cite{vincent2008extracting,burger2012image,mao2016image,TRND,zhang2017beyond}.
Inspired by U-Net~\cite{ronneberger2015u}, we propose a convolutional neural network for image denoising, which achieves the state-of-the-art performance. 

While popular image denoising algorithms reconstruct images 
by minimizing the
mean square error (MSE), important image details are usually lost which leads to image quality degradation, e.g., over-smoothing artifacts in some texture-rich regions are commonly observed in the denoised output from conventional methods, as shown in Fig.~\ref{fig:intro}(b). 
To this end, we propose a cascade architecture connecting image denoising to a high-level vision network.
We jointly minimize the image reconstruction loss 
and the high-level vision loss.
With the guidance of image semantic information, the denoising network is able to further improve visual quality and generate more visually appealing outputs, which demonstrates the importance of semantic information for image denoising.

When high-level vision tasks are conducted on noisy data, 
an independent image restoration step is typically applied as preprocessing, which is suboptimal for the ultimate goal~\cite{wang2016studying,wu2017relation,liu2017enhance}.
Recent research reveals that neural networks trained for image classification can be easily fooled by small noise perturbation or other artificial patterns~\cite{szegedy2013intriguing,nguyen2015deep}.
Therefore, an application-driven denoiser should be capable of simultaneously removing noise and preserving semantic-aware details for the high-level vision tasks.
Under the proposed architecture, we systematically investigate the mutual influence between the low-level and high-level vision networks.
We show that the cascaded network trained with the joint loss not only boosts the denoising network performance via image semantic guidance, but also substantially improves the accuracy of high-level vision tasks.
Moreover, our proposed training strategy makes the trained denoising network robust enough to different high-level vision tasks. 
In other words, our denoising module trained for one high-level vision task can be directly plugged into other high-level tasks without finetuning either module,
which facilitates the training effort when applied to various high-level tasks.

\section{Method}
\label{sec:method}

We first introduce the denoising network utilized in our framework, and then explain the relationship between the image denoising module and the module for high-level vision tasks in detail.

\subsection{Denoising Network}
\label{lab:denoising}


We propose a convolutional neural network for image denoising, which takes a noisy image as input and outputs the reconstructed image. 
This network conducts feature contraction and expansion through downsampling and upsampling operations, respectively. 
Each pair of downsampling and upsampling operations brings the feature representation into a new spatial scale, so that the whole network can process information on different scales.

Specifically, on each scale, the input is encoded  after downsampling the features from the previous scale.
After feature encoding and decoding possibly with features on the next scale, the output is upsampled and fused with the feature on the previous scale.
Such pairs of downsampling and upsampling steps can be nested to build deeper networks with more spatial scales of feature representation, which generally leads to better restoration performance.
Considering the tradeoff between computation cost and restoration accuracy, we choose three scales for the denoising network in our experiments, while this framework can be easily extended for more scales.

These operations together are designed to learn the residual between the input and the target output and recover as many details as possible, 
so we use a long-distance skip connection to sum the output of these operations and the input image, in order to generate the reconstructed image.
The overview is in Fig. \ref{fig:denoising} (a).
Each module in this network will be elaborated as follows.

\textbf{Feature Encoding}: We design one feature encoding module on each scale. which is one convolutional layer plus one residual block as in \cite{he2016deep}. 
The architecture is displayed in Fig.~\ref{fig:denoising} (b). 
Note that each convolutional layer is immediately followed by spatial batch normalization and a ReLU neuron. From top to down, the four convolutional layers have 128, 32, 32 and 128 kernels in size of $3 \times 3, 1 \times 1, 3 \times 3$ and $1 \times 1$, respectively.
The output of the first convolutioal layer is passed through a skip connection for element-wise sum with the output of the last convolutional layer.


\begin{figure*}[t]
	\center
	\includegraphics[width=0.7\linewidth]{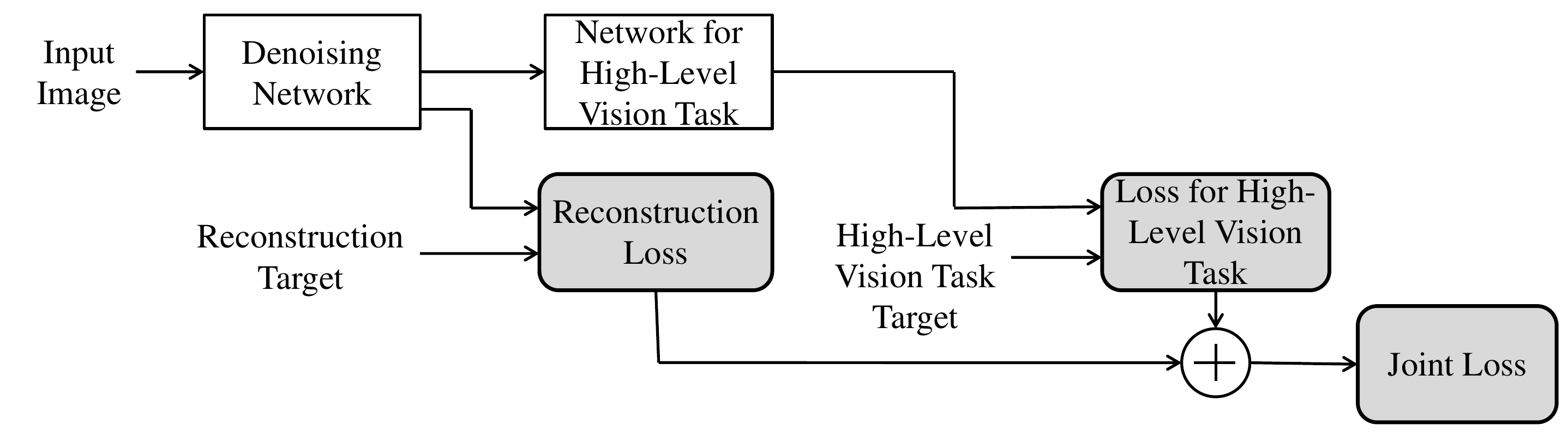}
	\vspace*{-0.1in}
	\caption{Overview of our proposed cascaded network.}
	\label{fig:system}
	\vspace*{-5mm}
\end{figure*}

\textbf{Feature Decoding}: The feature decoding module is designed for fusing information from two adjacent scales.
Two fusion schemes are tested: (1) concatenation of features on these two scales; (2) element-wise sum of them. 
Both of them obtain similar denosing performance.
Thus we choose the first scheme to accommodate feature representations of different channel numbers from two scales.
We use a similar architecture as the feature encoding module except that the number of kernels in the four convolutional layers are 256, 64, 64 and 256. 
Its architecture is in Fig.~\ref{fig:denoising}(c).

\textbf{Feature Downsampling \& Upsampling}: Downsampling operations are adopted multiple times to progressively increase the receptive field of the following convolution kernels and to reduce the computation cost by decreasing the feature map size.
The larger receptive field enables the kernels to incorporate 
larger spatial context for denoising.
We use 2 as the downsampling and upsampling factors, 
and try two schemes for downsampling in the experiments: (1) max pooling with stride of 2; (2) conducting convolutions with stride of 2. Both of them achieve similar denoisng performance in practice, so we use the second scheme in the rest experiments for computation efficiency. 
Upsampling operations are implemented by deconvolution with $4 \times 4$ kernels, which aim to expand the feature map to the same spatial size as the previous scale.

Since all the operations in our proposed denoising network are spatially invariant, it has the merit of handling input images of arbitrary size.


\subsection{When Image Denoising Meets High-Level Vision Tasks}

We propose a robust deep architecture processing a noisy image input, via cascading a  network for denoising and the other for high-level vision task, aiming to simultaneously: 
\vspace{-1mm}
\begin{enumerate}
	\setlength{\itemsep}{-2pt}
	\item reconstruct visually pleasing results guided by the high-level vision information, as the output of the denoisnig network;
	\item attain sufficiently good accuracy across various high-level vision tasks, when trained for only one high-level vision task;
\end{enumerate}
\vspace{-1mm}
The overview of the proposed cascaded network is displayed in Fig. \ref{fig:system}.
Specifically, given a noisy input image, the denosing network is first applied, and the denoised result is then fed into the following network for high-level vision task, which generates the high-level vision task output.

\textbf{Training Strategy}:
First we initialize the network for high-level vision task from a network that is well-trained in the noiseless setting.
We train the cascade of two networks in an end-to-end manner while
fixing the weights in the network for high-level vision task.
Only the weights in the denoising network are updated by the error back-propagated from the following network for high-level vision task,
which is similar to minimizing the perceptual loss for image super-resolution~\cite{johnson2016perceptual}.
The reason to adopt such a training strategy is to make the trained denoising network robust enough without losing the generality for various high-level vision tasks. 
More specifically, our denoising module trained for one high-level vision task can be directly plugged into other high-level tasks without finetuning either the denoiser or the high-level network.
Our approach not only facilitates the training effort when applying the denoiser to different high-level tasks while keeping the high-level vision network performing consistently for noisy and noise-free images, but also enables the denoising network to produce high-quality perceptual and semantically faithful results.

\textbf{Loss}:
The reconstruction loss of the denoising network is the mean squared error (MSE) between the denoising network output and the noiseless image. 
The losses of the classification network and the segmentation network both are the cross-entropy loss between the predicted label and the ground truth label.
The joint loss is defined as the weighted sum of the reconstruction loss and the loss for high-level vision task, which can be represented as
\begin{equation}
L(F(x), y) = L_D(F_D(x), \tilde{x}) + \lambda L_H(F_H(F_D(x)), y),
\label{eqn:loss}
\end{equation} 
\noindent
where $x$ is the noisy input image, $\tilde{x}$ is the noiseless image and $y$ is the ground truth label of high-level vision task. $F_D$, $F_H$ and $F$ denote the denoising network, the network of high-level vision task and the whole cascaded network, respectively.  $L_D$, $L_H$  represent the losses of the denoising network and the high-level vision task network, respectively, while $L$ is the joint loss, as illustrated in Fig.~\ref{fig:system}. $\lambda$ is the weight for balancing the losses $L_D$ and $L_H$.

\section{Experiments}

\begin{table*}[t]
	\centering
	\caption{
		Color image denoising results (PSNR) of different methods on Kodak dataset. The best result is shown in bold.
	}
	\vspace{-0.2cm}
	\label{tab:psnr} 
	\resizebox{\textwidth}{!}{
		\begin{tabular}{|@{\hskip 0.2mm}c@{\hskip 0.2mm}||c|c|@{\hskip 0.2mm}c@{\hskip 0.2mm}|c|c||c|@{\hskip 0.2mm}c@{\hskip 0.2mm}|c|c||c|c|@{\hskip 0.2mm}c@{\hskip 0.2mm}|c|c|}
			\hline
			& \multicolumn{5}{c||}{$\sigma = 25$} & \multicolumn{4}{c||}{$\sigma = 35$} & \multicolumn{5}{c|}{$\sigma = 50$} \\
			\hline
			\hline
			Image & \textbf{CBM3D} & \textbf{TNRD} & \textbf{MCWNNM} & \textbf{DnCNN} &  \textbf{Proposed}
			& \textbf{CBM3D} & \textbf{MCWNNM} & \textbf{DnCNN} &  \textbf{Proposed} 
			& \textbf{CBM3D} & \textbf{TNRD} & \textbf{MCWNNM} & \textbf{DnCNN} &  \textbf{Proposed} \\
			\hline
			\hline				
			01 
            & 29.13 & 27.21 & 28.66 & 29.75& \textbf{29.76}
			& 27.31 & 26.93 & 28.10 & \textbf{28.11}
			& 25.86 & 24.46 & 25.28 & 26.52 & \textbf{26.55} \\			
			\hline
			02 
            & 32.44 & 31.44 & 31.92 & 32.97& \textbf{33.00}
			& 31.07 & 30.62 & 31.65 & \textbf{31.75}
			& 29.84 & 29.12 & 29.27 & 30.44 & \textbf{30.54} \\	
			\hline
			03 
            & 34.54 & 32.73& 34.05 & 34.97& \textbf{35.12}
			& 32.62 & 32.27 & 33.37 & \textbf{33.58}
			& 31.34 & 29.95 & 30.52 & 31.76 & \textbf{31.99} \\			
			\hline
			04 
            & 32.67 & 31.16 & 32.42 & 32.94 & \textbf{33.01}
			& 31.02 & 30.92 & 31.51 & \textbf{31.59}
			& 29.92 & 28.65 & 29.37 & 30.12 & \textbf{30.22} \\
			\hline
			05 
            & 29.73 & 27.81 & 29.37 & 30.53& \textbf{30.55}
			& 27.61 & 27.53 & 28.66 & \textbf{28.72}
			& 25.92 & 24.37 & 25.60 & 26.77 & \textbf{26.87} \\			
			\hline
			06 
            & 30.59 & 28.52 & 30.18 & 31.05& \textbf{31.08}
			& 28.78 & 28.44 & 29.37 & \textbf{29.45}
			& 27.34 & 25.62 & 26.70 & 27.74 & \textbf{27.85} \\	
			\hline
			07 
            & 33.66 & 31.90 & 33.36 & 34.42& \textbf{34.47}
			& 31.64 & 31.53 & 32.60 & \textbf{32.70}
			& 29.99 & 28.24 & 29.51 & 30.67 & \textbf{30.82} \\			
			\hline
			08 
            & 29.88 & 27.38 & 29.39 & 30.30& \textbf{30.37}
			& 27.82 & 27.67 & 28.53 & \textbf{28.64}
			& 26.23 & 23.93 & 25.86 & 26.65 & \textbf{26.84} \\
			\hline
			09 
            & 34.06 & 32.21 & 33.42 & 34.59& \textbf{34.63}
			& 32.28 & 31.76 & 33.06 & \textbf{33.11}
			& 30.86 & 28.78 & 30.00 & 31.42 & \textbf{31.53} \\	
			\hline
			10 
            & 33.82 & 31.91 & 33.23 & 34.33& \textbf{34.38}
			& 31.97 & 31.51 & 32.74 & \textbf{32.83}
			& 30.48 & 28.78 & 29.63 & 31.03 & \textbf{31.17} \\			
			\hline
			11 
            & 31.25 & 29.51 & 30.62 & 31.82& \textbf{31.84}
			& 29.53 & 29.04 & 30.23 & \textbf{30.29}
			& 28.00 & 26.75 & 27.41 & 28.67 & \textbf{28.76} \\		
			\hline
			12 
            & 33.76 & 32.17 & 33.02 & 34.12& \textbf{34.18}
			& 32.24 & 31.52 & 32.73 & \textbf{32.83}
			& 30.98 & 29.70 & 30.00 & 31.32 & \textbf{31.47} \\
			\hline					
			13 
            & 27.64 & 25.52 & 27.19 & \textbf{28.26} & 28.24
			& 25.70 & 25.40 & 26.46 & \textbf{26.47}
			& 24.03 & 22.54 & 23.70 & 24.73 & \textbf{24.76} \\			
			\hline
			14 
            & 30.03 & 28.50 & 29.67 & 30.79 & \textbf{30.80}
			& 28.24 & 28.05 & 29.17 & \textbf{29.20}
			& 26.74 & 25.67 & 26.43 & 27.57 & \textbf{27.63} \\	
			\hline
			15 
            & 33.08 & 31.62 & 32.69 & 33.32 & \textbf{33.35}
			& 31.47 & 31.15 & 31.89 & \textbf{31.96}
			& 30.32 & 29.07 & 29.59 & 30.50 & \textbf{30.59} \\			
			\hline
			16 
            & 32.33 & 30.36 & 31.79 & 32.69 & \textbf{32.74}
			& 30.64 & 30.15 & 31.16 & \textbf{31.23}
			& 29.36 & 27.82 & 28.53 & 29.68 & \textbf{29.78} \\
			\hline
			17 
            & 32.93 & 31.20 & 32.39 & \textbf{33.53} & 33.50
			& 30.64 & 30.75 & 31.96 & \textbf{31.98}
			& 29.36 & 28.07 & 28.98 & 30.33 & \textbf{30.40} \\			
			\hline
			18 
            & 29.83 & 28.00 & 29.46 & 30.40& \textbf{30.46}
			& 28.00 & 27.70 & 28.72 & \textbf{28.79}
			& 26.41 & 25.06 & 25.94 & 27.03 & \textbf{27.14} \\	
			\hline
			19 
            & 31.78 & 30.01 & 31.29 & \textbf{32.23} & 32.30
			& 30.19 & 29.86 & 30.80 & \textbf{30.88}
			& 29.06 & 27.30 & 28.44 & 29.34 & \textbf{29.49} \\			
			\hline
			20 
            & 33.45 & 32.00 & 32.78 & 34.15& \textbf{34.29}
			& 31.84 & 31.32 & 32.73 & \textbf{32.91}
			& 30.51 & 29.24 & 29.79 & 31.28 & \textbf{31.53} \\
			\hline
			21 
            & 30.99 & 29.09 & 30.55 & 31.61 & \textbf{31.63}
			& 29.17 & 28.86 & 29.94 & \textbf{29.98}
			& 27.61 & 26.09 & 27.13 & 28.27 & \textbf{28.34} \\	
			\hline
			22 
            & 30.93 & 29.60 & 30.48 & \textbf{31.41} & 31.38
			& 29.36 & 28.93 & 29.94 & \textbf{29.95}
			& 28.09 & 27.14 & 27.47 & 28.54 & \textbf{28.58} \\			
			\hline
			23 
            & 34.79 & 33.68 & 34.45 & 35.36 & \textbf{35.40}
			& 33.09 & 32.79 & 33.86 & \textbf{33.89}
			& 31.75 & 30.53 & 30.96 & 32.18 & \textbf{32.30} \\		
			\hline
			24 
            & 30.09 & 28.17  & 29.93 & \textbf{30.79} & 30.77
			& 28.19 & 28.17 & 28.98 & \textbf{29.03}
			& 26.62 & 24.92 & 26.37 & 27.18 & \textbf{27.30} \\
			\hline	
			\hline				
			\textbf{Average} 
            & 31.81 & 30.08 & 31.35 & 32.35 & \textbf{32.39}
			& 30.04 & 29.70 & 30.76 & \textbf{30.83}
			& 28.62 & 27.17 & 28.02 & 29.16 & \textbf{29.27} \\																	
			\hline
		\end{tabular}
	}
	\vspace{-4mm}
\end{table*}

\subsection{Image Denoising}

Our proposed denoising network takes RGB images as input,  
and outputs the reconstructed images directly.
We add independent and identically distributed Gaussian noise with zero mean to the original image as the noisy input image during training.
We use the training set as in~\cite{chen2014semantic}. 
The loss of training is equivalent to Eqn.~\ref{eqn:loss} as $\lambda = 0$.
We use SGD with a batch size of 32, and the input patches are $48 \times 48$ pixels.
The initial learning rate is set as $10^{-4}$ and is divided by 10 after every 500,000 iterations. The training is terminated after 1,500,000 iterations.
We train a different denoising network for each noise level in our experiment.




\begin{figure*}[t]
	\center
	\begin{tabular}{@{\hskip 0.5mm}c@{\hskip 0.5mm}c@{\hskip 1mm}c@{\hskip 0.5mm}c}
		\includegraphics[height=0.19\linewidth, trim=0 0 2 309,clip]{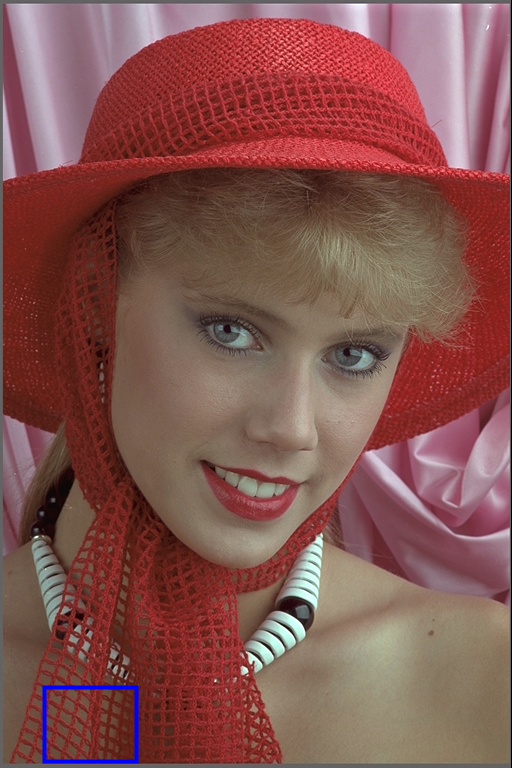} &
		\includegraphics[height=0.19\linewidth, trim=45 8 387 688,clip]{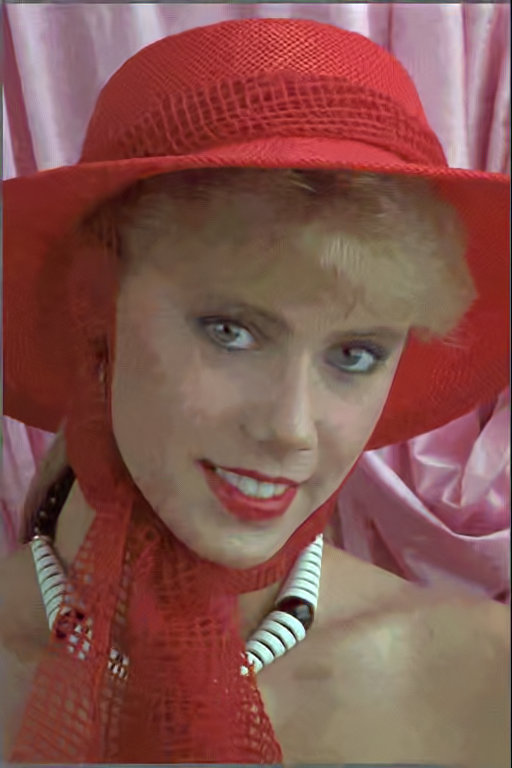} &
		\includegraphics[height=0.19\linewidth, trim=88 0 40 32,clip]{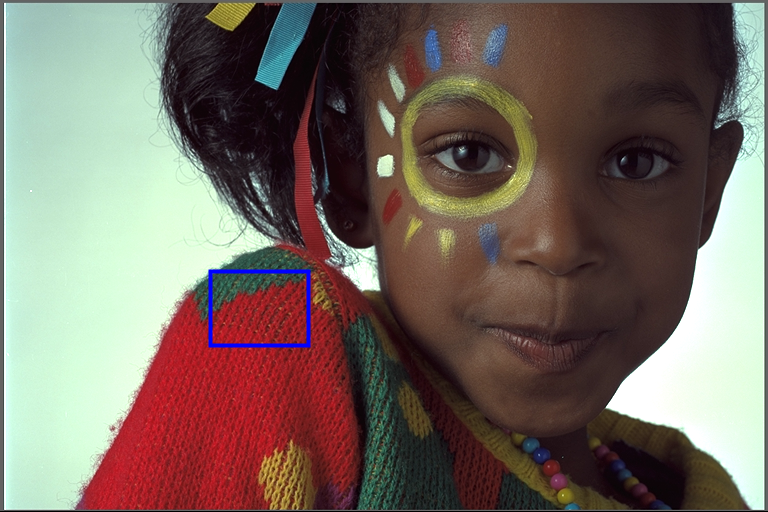} &
		\includegraphics[height=0.19\linewidth, trim=211 168 461 272,clip]{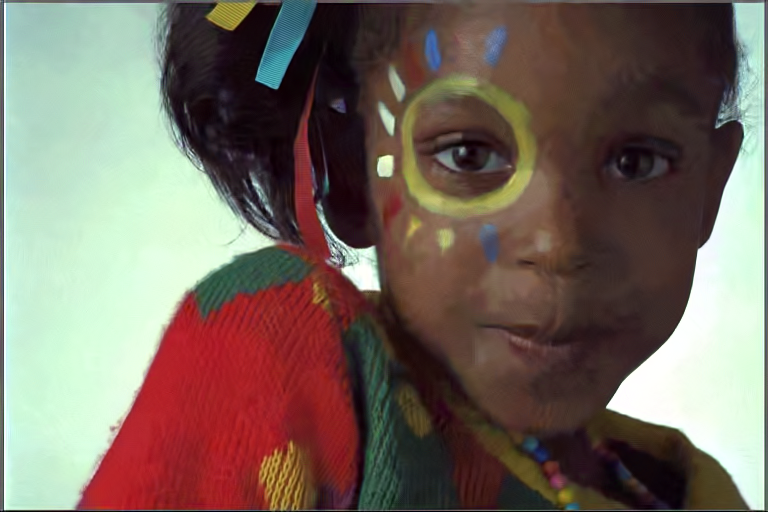}  \\
		 (I)  & (II)  & (I)  & (II)  \\
		\includegraphics[height=0.19\linewidth, trim=45 8 387 688,clip]{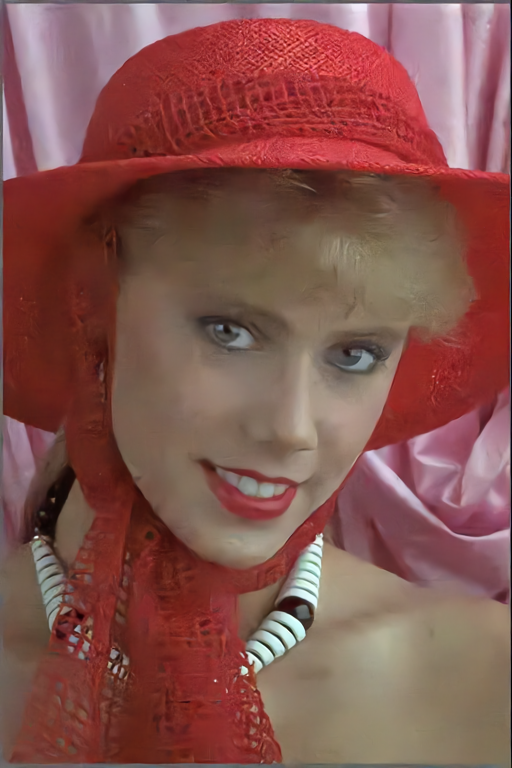} &
		\includegraphics[height=0.19\linewidth, trim=45 8 387 688,clip]{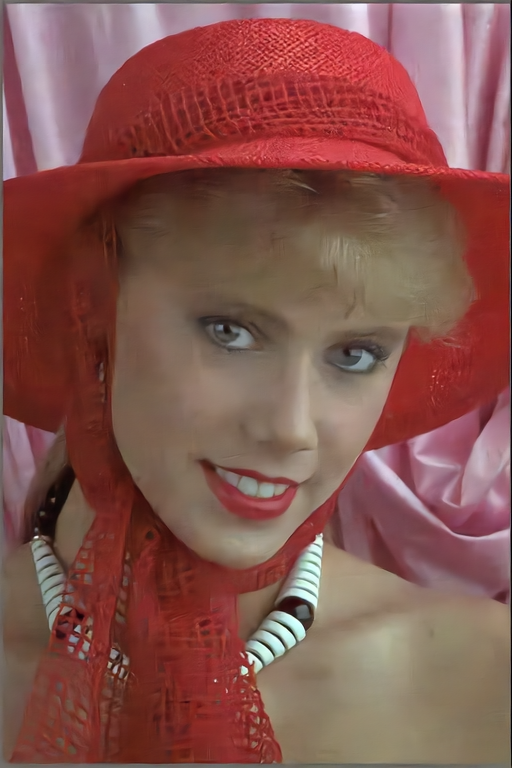} &		
		\includegraphics[height=0.19\linewidth, trim=211 168 461 272,clip]{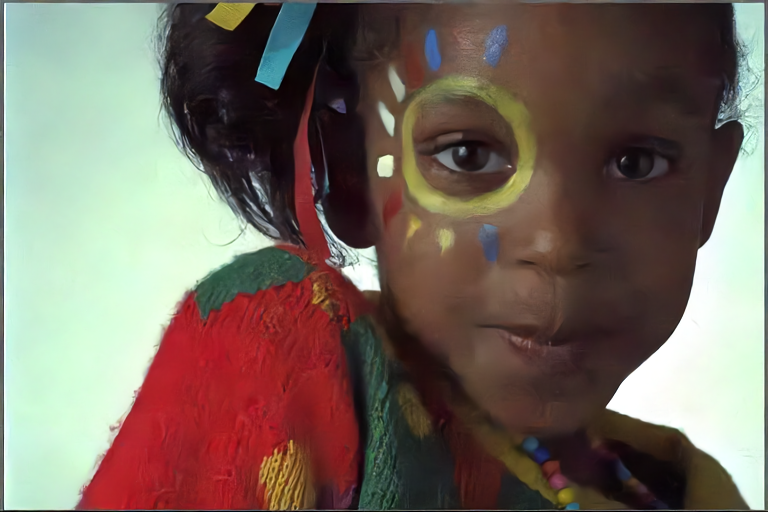} &
		\includegraphics[height=0.19\linewidth, trim=211 168 461 272,clip]{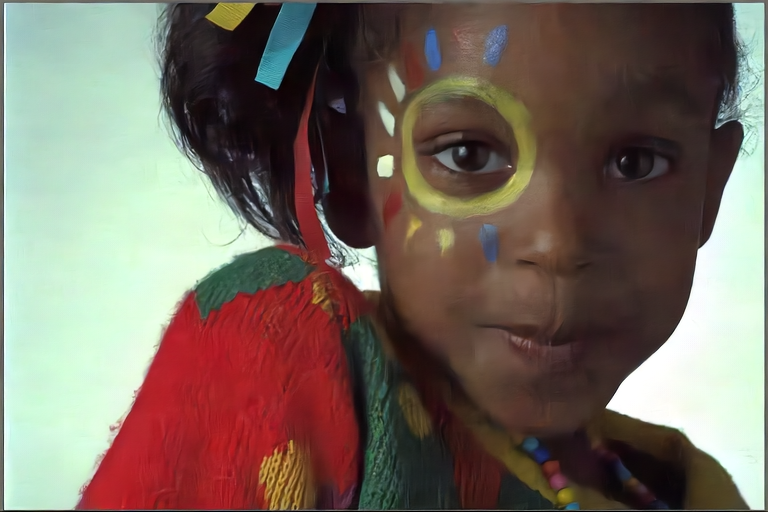} \\
		(III)  & (IV)  & (III)  & (IV) \\	
		\includegraphics[height=0.19\linewidth, trim=45 8 387 688,clip]{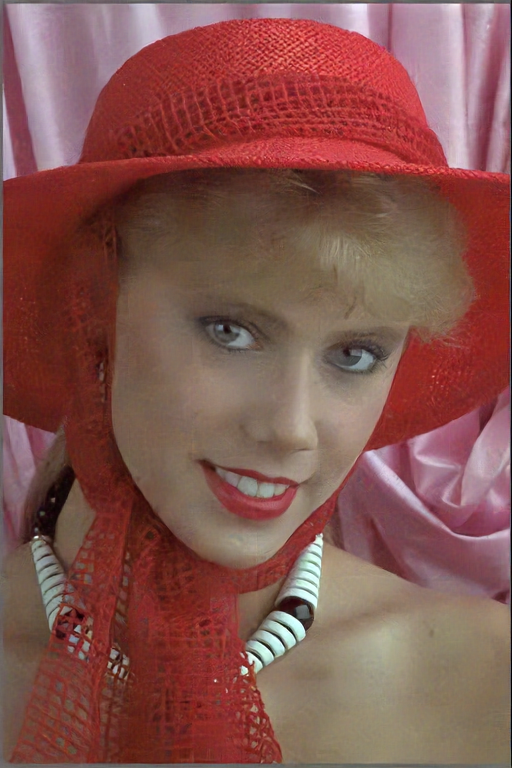}	&
		\includegraphics[height=0.19\linewidth, trim=45 8 387 688,clip]{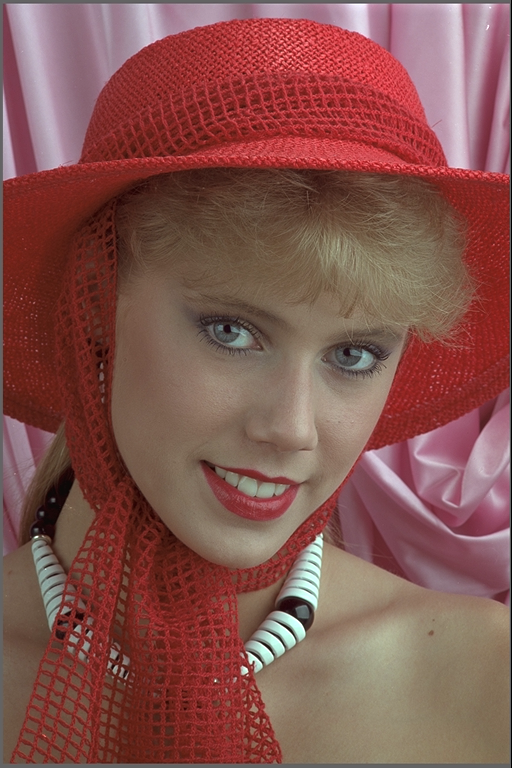} &		
		\includegraphics[height=0.19\linewidth, trim=211 168 461 272,clip]{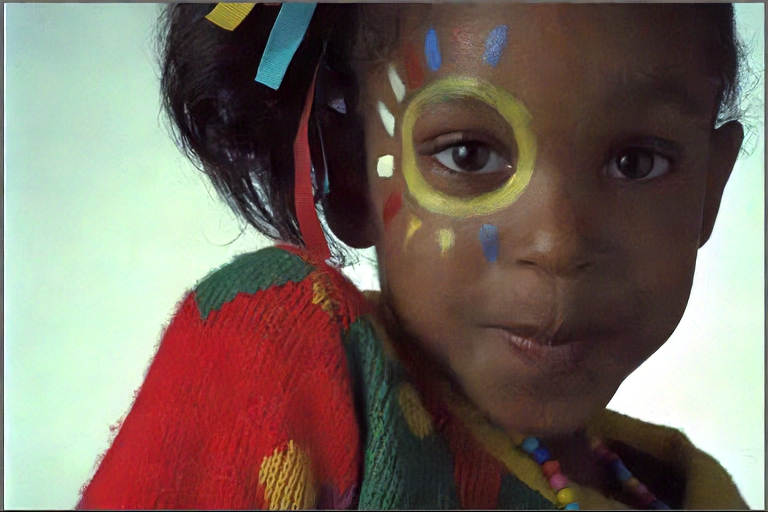}  &
		\includegraphics[height=0.19\linewidth, trim=211 168 461 272,clip]{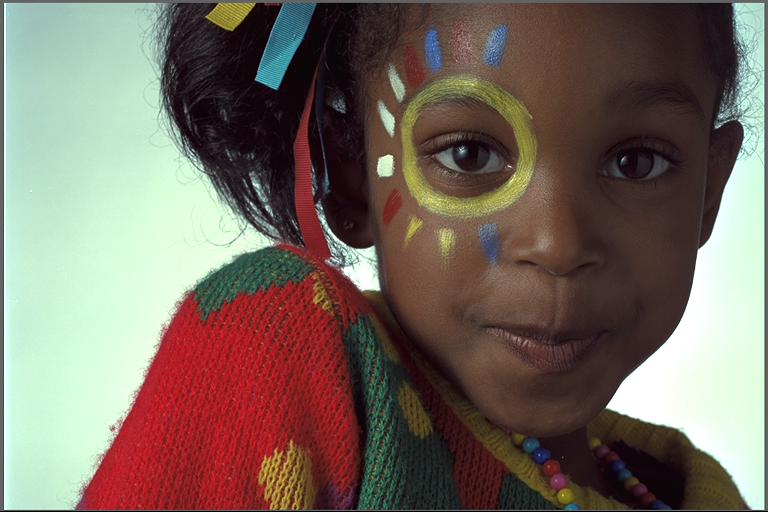} \\
		(V)  & (VI) &  (V)  & (VI) \\
		\multicolumn{2}{c}{(a)} & \multicolumn{2}{c}{(b)} \\
	\end{tabular}
	\vspace{-4mm}
	\caption{(a) Two image denoising examples from Kodak dataset. We show (I) the ground truth image and the zoom-in regions of: (II) the denoised image by CBM3D; (III) the denoised image by DnCNN; the denoising result of our proposed model (IV) without the guidance of high-level vision information; (V) with the guidance of high-level vision information and (VI) the ground truth. }
	\vspace*{-4mm}
	\label{fig:denoise3}
\end{figure*}

We compare our denoisnig network with several state-of-the-art color image denoising approaches on various noise levels: $\sigma = 25, 35$ and $50$.
We evaluate their denoising performance
over the widely used Kodak dataset\footnote{\url{http://r0k.us/graphics/kodak/}}, which consists of 24 color images.
Table~\ref{tab:psnr} shows the peak signal-to-noise ratio (PSNR) results for CBM3D \cite{dabov2007color}, TNRD~\cite{TRND}, MCWNNM~\cite{xu2017multi}, DnCNN~\cite{zhang2017beyond}, and our proposed method. 
We do not list other methods \cite{burger2012image,zoran2011learning,WNNM,zhang2017learning} whose average performance is wore than DnCNN.
The implementation codes used are from the authors’ websites and the default parameter settings are adopted in our experiments.\footnote{For TNRD, we denoise each color channel using the grayscale image denoising implementation which is from the authors' website. TNRD for $\sigma=35$ is not publicly available, so we do not include this case here.}


It is clear that our proposed method outperforms all the competing approaches quantitatively across different noise levels.
It achieves the highest PSNR in almost every image of Kodak dataset.
 


\subsection{When Image Denoising Meets High-Level Vision Tasks}

We choose two high-level vision tasks as representatives in our study: image classification and semantic segmentation, which have been dominated by deep network based models. 
We utilize two popular VGG-based deep networks in our system for each task, respectively.
\textit{VGG-16} in~\cite{simonyan2014very} is employed for image classification;
we select \textit{DeepLab-LargeFOV} in~\cite{chen2014semantic} for semantic segmentation.
We follow the preprocessing protocols (e.g. crop size, mean removal of each color channel) in~\cite{simonyan2014very} and~\cite{chen2014semantic} accordingly while training and deploying them in our experiments.

As for the cascaded network for image classification and the corresponding experiments, we train our model on ILSVRC2012 training set, 
and evaluate the classification accuracy on ILSVRC2012 validation set. 
$\lambda$ is empirically set as $0.25$.
As for the cascaded network for image semantic segmentation and its corresponding experiments, we train our model on the augmented training set of Pascal VOC 2012 as in~\cite{chen2014semantic}, and test on its validation set. 
$\lambda$ is empirically set as $0.5$.

\subsubsection{High-Level Vision Information Guided Image Denoising}

The 
typical
metric used for image denoising is PSNR, which has been shown to 
sometimes 
correlate poorly with human assessment of visual quality~\cite{huynh2008scope}. 
Since PSNR depends on the reconstruction error between the denoised output and the reference image, a model trained by minimizing MSE on the image domain should always outperform a model trained by minimizing our proposed joint loss (with the guidance of high-level vision semantics) in the metric of PSNR.
Therefore, we emphasize that the goal of our following experiments is not to pursue the highest PSNR, but to demonstrate the qualitative difference between the model trained with our proposed joint loss and the model trained with MSE on the image domain.

Fig.~\ref{fig:denoise3} displays two image denoising examples from Kodak dataset. 
A visual comparison is illustrated for a zoom-in region: (II) and (III) are the denoising results using CBM3D~\cite{dabov2007color}, and DnCNN~\cite{zhang2017beyond}, respectively; (IV) is the proposed denoiser trained separately without the guidance of high-level vision information; (V) is the denoising result using the proposed denoising network trained jointly with a segmentation network.
We can find that the results using CBM3D, DnCNN and our separately trained denoiser generate oversmoothing regions, while the jointly trained denoising network is able to reconstruct the denoised image which preserves more details and textures with better visual quality. 

\subsubsection{Generality of the Denoiser for High-Level Vision Tasks}
\label{exp_high_level}

\begin{figure*}[t!]
\begin{center}
\begin{tabular}{c@{\hskip 0.5mm}c@{\hskip 0.5mm}c@{\hskip 0.5mm}c}
\includegraphics[width=0.23\linewidth, trim=0 0 0 36, clip]{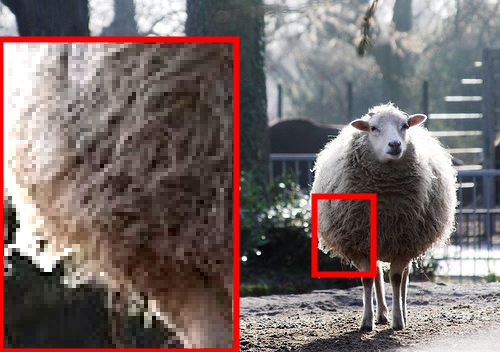} &
\includegraphics[width=0.23\linewidth, trim=0 0 0 36, clip]{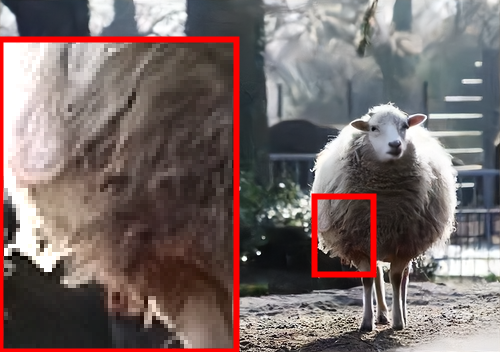} &
\includegraphics[width=0.23\linewidth, trim=0 0 0 36, clip]{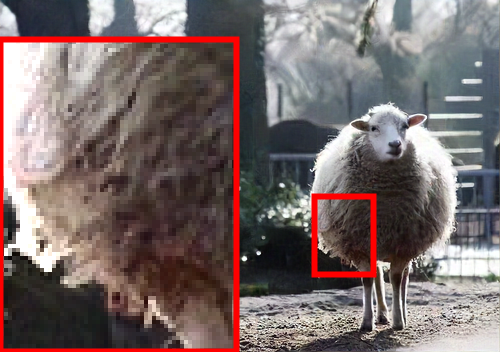} & 
\includegraphics[width=0.23\linewidth, trim=0 0 0 36, clip]{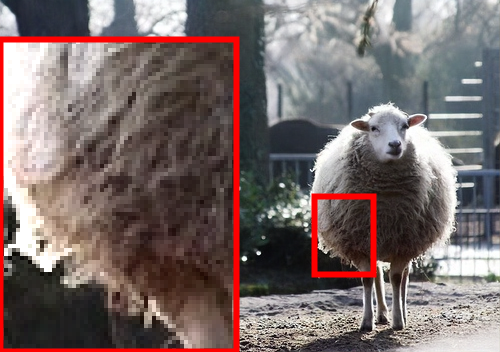} \\
\includegraphics[width=0.23\linewidth, trim=0 0 0 36, clip]{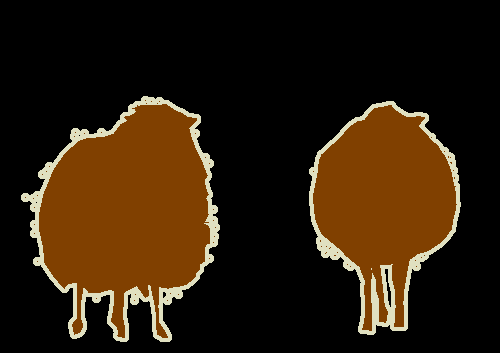} &
\includegraphics[width=0.23\linewidth, trim=0 0 0 36, clip]{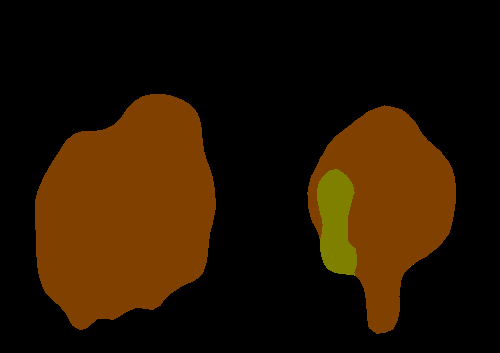} &
\includegraphics[width=0.23\linewidth, trim=0 0 0 36, clip]{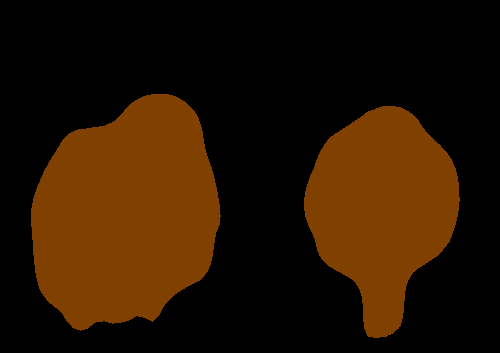} & 
\includegraphics[width=0.23\linewidth, trim=0 0 0 36, clip]{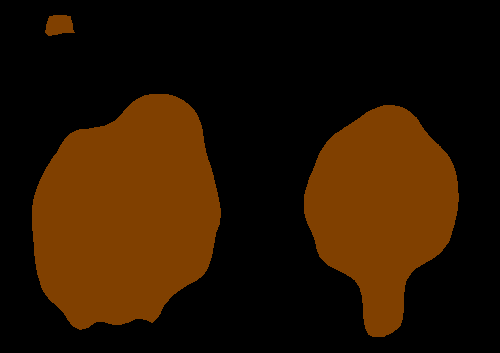} \\

\includegraphics[width=0.23\linewidth]{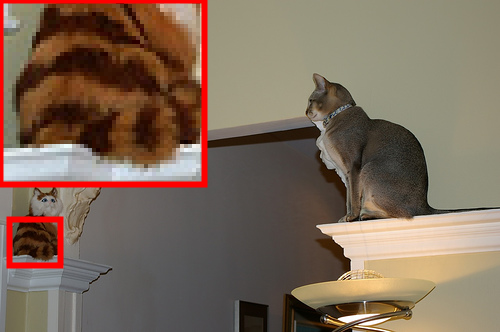} &
\includegraphics[width=0.23\linewidth]{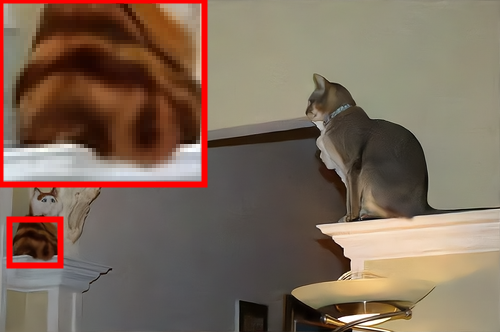} &
\includegraphics[width=0.23\linewidth]{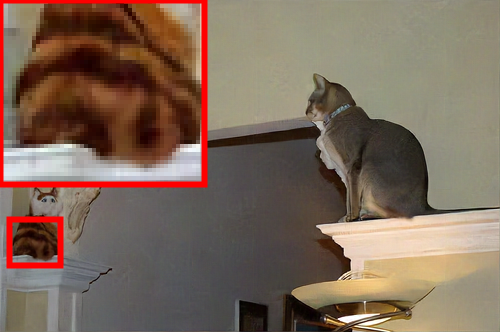} & 
\includegraphics[width=0.23\linewidth]{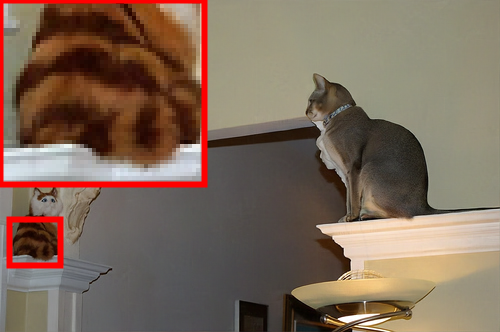} \\
\includegraphics[width=0.23\linewidth]{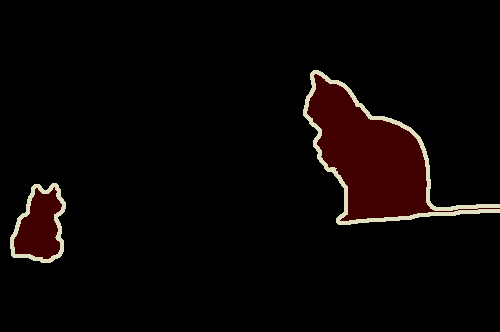} &
\includegraphics[width=0.23\linewidth]{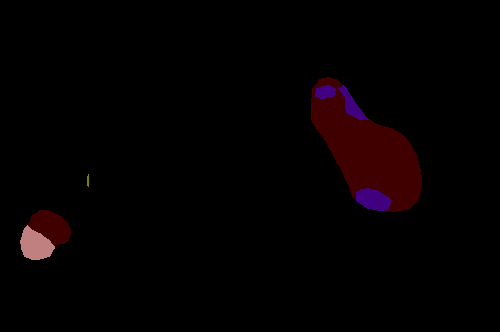} &
\includegraphics[width=0.23\linewidth]{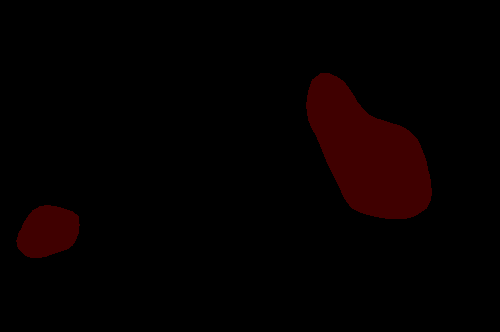} & 
\includegraphics[width=0.23\linewidth]{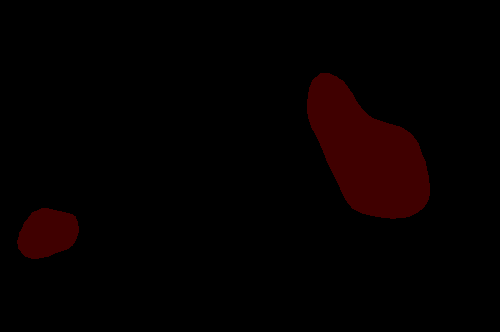} \\
{\small (a)} & {\small (b)} & {\small (c)} & {\small (d)} \\
\end{tabular}
\end{center}
\vspace{-4mm}
\caption{Two semantic segmentation examples from Pascal VOC 2012 validation set. From left to right: (a) the ground truth image, the denoised image using (b) the separately trained denoiser, (c) the denoiser trained with the reconstruction and segmentation joint loss, and (d) the denoiser trained with the classification network and evaluated for semantic segmentation. 
Their corresponding segmentation label maps are shown below. The zoom-in region which generates inaccurate segmentation in (b) is displayed in the red box.
}
\label{fig:seg1}
\vspace{-5mm}
\end{figure*}

We now investigate how the image denoising can enhance the high-level vision applications, including image classification and semantic segmentation, over the ILSVRC2012 and Pascal VOC 2012 datasets, respectively.
The noisy images ($\sigma = 15, 30, 45, 60$) are denoised and then fed into the VGG-based networks for high-level vision tasks. 
To evaluate how different denoising schemes contribute to the  performance of high-level vision tasks, we experiment with the following cases:
\vspace{-1mm}

\begin{table}
	\begin{center}

		\fontsize{8}{10pt}\selectfont
		\caption{Classification accuracy after denoising noisy image input, averaged over ILSVRC2012 validation dataset. \textcolor{red}{Red} is the best and \textcolor{blue}{blue} is the second best results. }
		\label{table:classDenoising1}		
		\vspace{-2mm}
		\begin{tabular}{|@{\hskip 1mm}c@{\hskip 1mm}|@{\hskip 1mm}c@{\hskip 1mm}|c|@{\hskip 1mm}c@{\hskip 1mm}|@{\hskip 1mm}c@{\hskip 1mm}|@{\hskip 1mm}c@{\hskip 1mm}|@{\hskip 1mm}c@{\hskip 1mm}|}		
			\hline
			\multicolumn{2}{|c|}{} & VGG & CBM3D + & Separate + & Joint & Joint Training\\   [-0.3ex] 
			\multicolumn{2}{|c|}{} & & VGG & VGG & Training & (Cross-Task) \\   
			\hline 
			\hline
			\multirow{2}{*}{$\sigma$=15} 
			& Top-1 & 62.4 & 68.2 & 68.3 & \textcolor{red}{69.9} & \textcolor{blue}{69.8} \\ [-0.3ex]
			& Top-5 & 84.2 & 88.8 & 88.7 & \textcolor{red}{89.5} & \textcolor{blue}{89.4} \\ 
			\hline
			\multirow{2}{*}{$\sigma$=30} 
			& Top-1 & 44.4 & 62.3 & 62.7 & \textcolor{red}{67.0} & \textcolor{blue}{66.4} \\ [-0.3ex]
			& Top-5 & 68.9 & 84.8 & 84.9 & \textcolor{red}{87.6} & \textcolor{blue}{87.2} \\ 
			\hline
			\multirow{2}{*}{$\sigma$=45} 
			& Top-1 & 24.3 & 55.2 & 54.6 & \textcolor{red}{63.0} & \textcolor{blue}{62.0} \\ [-0.3ex]
			& Top-5 & 46.1 & 79.4 & 78.8 & \textcolor{red}{84.6} & \textcolor{blue}{84.0} \\ 
			\hline
			\multirow{2}{*}{$\sigma$=60} 
			& Top-1 & 11.4 & 50.0 & 50.1 & \textcolor{red}{59.2} & \textcolor{blue}{57.0} \\ [-0.3ex]
			& Top-5 & 26.3 & 74.2 & 74.5 & \textcolor{red}{81.8} & \textcolor{blue}{80.2} \\ 
			\hline	
		\end{tabular}
		
	\end{center}
	\vspace{-5mm}
\end{table}

\begin{itemize}
	\setlength{\itemsep}{-2pt}
	\item noisy images are directly fed into the high-level vision network, termed as \textit{VGG}. This approach serves as the baseline;
	\item noisy images are first denoised by CBM3D, and then fed into the high-level vision network, termed as \textit{CBM3D+VGG};
	\item noisy images are denoised via the separately trained denoising network, and then fed into the high-level vision network, termed as \textit{Separate+VGG};
	\item our proposed approach: noisy images are processed by the cascade of these two networks, which is trained using the joint loss, termed as \textit{Joint Training}.
	\item a denoising network is trained with the classification network in our proposed approach, but then is connected to the segmentation network and evaluated for the task of semantic segmentation, or vice versa. This is to validate the generality of our denoiser for various high-level tasks, termed as \textit{Joint Training (Cross-Task).
}
\end{itemize}
\vspace{-1mm}
Note that the weights in the high-level vision network are initialized from a well-trained network under the noiseless setting and not updated during training in our experiments.


\begin{table}
	\begin{center}
		\fontsize{8}{10pt}\selectfont
		\caption{Segmentation results (mIoU) after denoising noisy image input, averaged over  Pascal VOC 2012 validation dataset. \textcolor{red}{Red} is the best and \textcolor{blue}{blue} is the second best results.}	
		\label{table:classDenoising2}				
		\vspace{-2mm}	
		\begin{tabular}{|c|c|@{\hskip 1mm}c@{\hskip 1mm}|@{\hskip 1mm}c@{\hskip 1mm}|@{\hskip 1mm}c@{\hskip 1mm}|@{\hskip 1mm}c@{\hskip 1mm}|}		
			\hline
			 & VGG & CBM3D + & Separate + & Joint & Joint Training \\   [-0.3ex] 
			 & & VGG & VGG &  Training & (Cross-Task) \\   
			\hline 
			\hline
			
			$\sigma$=15 & 56.78 & 59.58 & 58.70 & \textcolor{red}{60.46} & \textcolor{blue}{60.41} \\
			\hline	
			
			$\sigma$=30  & 43.43 & 55.29 & 54.13 & \textcolor{red}{57.86} & \textcolor{blue}{56.29} \\
			\hline
			
			$\sigma$=45  & 27.99 & 50.69 & 49.51 & \textcolor{red}{54.83} & \textcolor{blue}{54.01} \\
			\hline
			
			$\sigma$=60  & 14.94 & 46.56 & 46.59 & \textcolor{red}{52.02} & \textcolor{blue}{51.82} \\
			\hline			
		\end{tabular}
		
	\end{center}
	\vspace{-5mm}
\end{table}

Table~\ref{table:classDenoising1} and Table~\ref{table:classDenoising2} list the performance of high-level vision tasks, i.e., top-1 and top5 accuracy for classification and mean intersection-over-union (IoU) without conditional random field (CRF) postprocessing for semantic segmentation. 
We notice that the baseline VGG approach obtains much lower accuracy than all the other cases, which shows the necessity of image denoising as a preprocessing step for high-level vision tasks on noisy data.
When we only apply denoising without considering high-level semantics (e.g., in CBM3D+VGG and Separate+VGG), it also fails to achieve high accuracy due to the artifacts introduced by the denoisers.
The proposed Joint Training approach achieves sufficiently high accuracy across various noise levels.


As for the case of Joint Training (Cross-Task), first we train the denoising network jointly with the segmentation network and then connect this denoiser to the classification network.
As shown in Table~\ref{table:classDenoising1},  
its accuracy remarkably outperforms the cascade of a separately trained denoising network and a classification network (i.e., Separate+VGG), and is comparable to our proposed model dedicatedly trained for classification (Joint Training). 
In addition, we use the denoising network jointly trained with the classification network, to connect the segmentation network. Its mean IoU is much better than Separate+VGG in Table~\ref{table:classDenoising2}. 
These two experiments show the high-level semantics of different tasks are universal in terms of low-level vision tasks, which is in line with intuition, and the denoiser trained in our method has the generality for various high-level tasks.

Fig.~\ref{fig:seg1} displays two visual examples of how the data-driven denoising can enhance the semantic segmentation performance. 
It is observed that the segmentation result of the denoised image from the separately trained denoising network has lower accuracy compared to those using the joint loss and the joint loss (cross-task), while the zoom-in region of its denoised image for inaccurate segmentation in Fig.~\ref{fig:seg1}~(b) contains oversmoothing artifacts.
On the contrary, both the Joint Training and Joint Training (Cross-Task) approaches achieve finer segmentation result and produce more visually pleasing denoised outputs simultaneously. 

\section{Conclusion}
Exploring the connection between low-level vision and high-level semantic tasks is of great practical value in various applications of computer vision.
In this paper, we tackle this challenge in a simple yet efficient way by allowing the high-level semantic information flowing back to the low-level vision part,
which achieves superior performance in both image denoising and various high-level vision tasks.
In our method, the denoiser trained for one high-level task has the generality to other high-level vision tasks.
Overall, it provides a feasible and robust solution in a deep learning fashion to real world problems,
which can be used to handle other corruptions~\cite{liu2016robust,li2017aod}.

\bibliographystyle{named}
\bibliography{ijcai18}

\begin{thebibliography}{}

\bibitem[\protect\citeauthoryear{Aharon \bgroup \em et al.\egroup
  }{2006}]{Aharon2006}
Michal Aharon, Michael Elad, and Alfred Bruckstein.
\newblock {K-SVD} : An algorithm for designing overcomplete dictionaries for
  sparse representation.
\newblock {\em TSP}, 54(11):4311--4322, 2006.

\bibitem[\protect\citeauthoryear{Burger \bgroup \em et al.\egroup
  }{2012}]{burger2012image}
Harold~C Burger, Christian~J Schuler, and Stefan Harmeling.
\newblock Image denoising: Can plain neural networks compete with bm3d?
\newblock In {\em CVPR}, pages 2392--2399. IEEE, 2012.

\bibitem[\protect\citeauthoryear{Chen and Pock}{2017}]{TRND}
Yunjin Chen and Thomas Pock.
\newblock Trainable nonlinear reaction diffusion: A flexible framework for fast
  and effective image restoration.
\newblock {\em TPAMI}, 39(6):1256--1272, 2017.

\bibitem[\protect\citeauthoryear{Chen \bgroup \em et al.\egroup
  }{2014}]{chen2014semantic}
Liang-Chieh Chen, George Papandreou, Iasonas Kokkinos, Kevin Murphy, and Alan~L
  Yuille.
\newblock Semantic image segmentation with deep convolutional nets and fully
  connected crfs.
\newblock {\em arXiv preprint arXiv:1412.7062}, 2014.

\bibitem[\protect\citeauthoryear{Dabov \bgroup \em et al.\egroup
  }{2007a}]{dabov2007color}
Kostadin Dabov, Alessandro Foi, Vladimir Katkovnik, and Karen Egiazarian.
\newblock Color image denoising via sparse 3d collaborative filtering with
  grouping constraint in luminance-chrominance space.
\newblock In {\em ICIP}, volume~1, pages I--313. IEEE, 2007.

\bibitem[\protect\citeauthoryear{Dabov \bgroup \em et al.\egroup
  }{2007b}]{dabov2007image}
Kostadin Dabov, Alessandro Foi, Vladimir Katkovnik, and Karen Egiazarian.
\newblock Image denoising by sparse 3-d transform-domain collaborative
  filtering.
\newblock {\em TIP}, 16(8):2080--2095, 2007.

\bibitem[\protect\citeauthoryear{Dong \bgroup \em et al.\egroup }{2013}]{NCSR}
Weisheng Dong, Lei Zhang, Guangming Shi, and Xin Li.
\newblock Nonlocally centralized sparse representation for image restoration.
\newblock {\em TIP}, 22(4):1620--1630, 2013.

\bibitem[\protect\citeauthoryear{Gu \bgroup \em et al.\egroup }{2014}]{WNNM}
Shuhang Gu, Lei Zhang, Wangmeng Zuo, and Xiangchu Feng.
\newblock Weighted nuclear norm minimization with application to image
  denoising.
\newblock In {\em CVPR}, pages 2862--2869, 2014.

\bibitem[\protect\citeauthoryear{He \bgroup \em et al.\egroup
  }{2016}]{he2016deep}
Kaiming He, Xiangyu Zhang, Shaoqing Ren, and Jian Sun.
\newblock Deep residual learning for image recognition.
\newblock In {\em CVPR}, pages 770--778, 2016.

\bibitem[\protect\citeauthoryear{Huynh-Thu and Ghanbari}{2008}]{huynh2008scope}
Quan Huynh-Thu and Mohammed Ghanbari.
\newblock Scope of validity of psnr in image/video quality assessment.
\newblock {\em Electronics letters}, 44(13):800--801, 2008.

\bibitem[\protect\citeauthoryear{Johnson \bgroup \em et al.\egroup
  }{2016}]{johnson2016perceptual}
Justin Johnson, Alexandre Alahi, and Li~Fei-Fei.
\newblock Perceptual losses for real-time style transfer and super-resolution.
\newblock In {\em ECCV}, pages 694--711. Springer, 2016.

\bibitem[\protect\citeauthoryear{Li \bgroup \em et al.\egroup
  }{2017}]{li2017aod}
Boyi Li, Xiulian Peng, Zhangyang Wang, Jizheng Xu, and Dan Feng.
\newblock Aod-net: All-in-one dehazing network.
\newblock In {\em ICCV}, 2017.

\bibitem[\protect\citeauthoryear{Liu \bgroup \em et al.\egroup
  }{2016}]{liu2016robust}
Ding Liu, Zhaowen Wang, Bihan Wen, Jianchao Yang, Wei Han, and Thomas~S Huang.
\newblock Robust single image super-resolution via deep networks with sparse
  prior.
\newblock {\em IEEE TIP}, 25(7):3194--3207, 2016.

\bibitem[\protect\citeauthoryear{Liu \bgroup \em et al.\egroup
  }{2017}]{liu2017enhance}
Ding Liu, Bowen Cheng, Zhangyang Wang, Haichao Zhang, and Thomas~S Huang.
\newblock Enhance visual recognition under adverse conditions via deep
  networks.
\newblock {\em arXiv preprint arXiv:1712.07732}, 2017.

\bibitem[\protect\citeauthoryear{Mairal \bgroup \em et al.\egroup
  }{2009}]{Mairal2009}
J.~Mairal, F.~Bach, J.~Ponce, G.~Sapiro, and A.~Zisserman.
\newblock Non-local sparse models for image restoration.
\newblock In {\em ICCV}, 2009.

\bibitem[\protect\citeauthoryear{Mao \bgroup \em et al.\egroup
  }{2016}]{mao2016image}
Xiaojiao Mao, Chunhua Shen, and Yu-Bin Yang.
\newblock Image restoration using very deep convolutional encoder-decoder
  networks with symmetric skip connections.
\newblock In {\em NIPS}, pages 2802--2810. 2016.

\bibitem[\protect\citeauthoryear{Nguyen \bgroup \em et al.\egroup
  }{2015}]{nguyen2015deep}
Anh Nguyen, Jason Yosinski, and Jeff Clune.
\newblock Deep neural networks are easily fooled: High confidence predictions
  for unrecognizable images.
\newblock In {\em CVPR}, pages 427--436, 2015.

\bibitem[\protect\citeauthoryear{Ronneberger \bgroup \em et al.\egroup
  }{2015}]{ronneberger2015u}
Olaf Ronneberger, Philipp Fischer, and Thomas Brox.
\newblock U-net: Convolutional networks for biomedical image segmentation.
\newblock In {\em MICCAI}, pages 234--241. Springer, 2015.

\bibitem[\protect\citeauthoryear{Simonyan and
  Zisserman}{2014}]{simonyan2014very}
Karen Simonyan and Andrew Zisserman.
\newblock Very deep convolutional networks for large-scale image recognition.
\newblock {\em arXiv preprint arXiv:1409.1556}, 2014.

\bibitem[\protect\citeauthoryear{Szegedy \bgroup \em et al.\egroup
  }{2013}]{szegedy2013intriguing}
Christian Szegedy, Wojciech Zaremba, Ilya Sutskever, Joan Bruna, Dumitru Erhan,
  Ian Goodfellow, and Rob Fergus.
\newblock Intriguing properties of neural networks.
\newblock {\em arXiv preprint arXiv:1312.6199}, 2013.

\bibitem[\protect\citeauthoryear{Vincent \bgroup \em et al.\egroup
  }{2008}]{vincent2008extracting}
Pascal Vincent, Hugo Larochelle, Yoshua Bengio, and Pierre-Antoine Manzagol.
\newblock Extracting and composing robust features with denoising autoencoders.
\newblock In {\em ICML}, pages 1096--1103. ACM, 2008.

\bibitem[\protect\citeauthoryear{Wang \bgroup \em et al.\egroup
  }{2016}]{wang2016studying}
Zhangyang Wang, Shiyu Chang, Yingzhen Yang, Ding Liu, and Thomas~S Huang.
\newblock Studying very low resolution recognition using deep networks.
\newblock In {\em CVPR}, pages 4792--4800, 2016.

\bibitem[\protect\citeauthoryear{Wu \bgroup \em et al.\egroup
  }{2017}]{wu2017relation}
Jiqing Wu, Radu Timofte, Zhiwu Huang, and Luc Van~Gool.
\newblock On the relation between color image denoising and classification.
\newblock {\em arXiv preprint arXiv:1704.01372}, 2017.

\bibitem[\protect\citeauthoryear{Xu \bgroup \em et al.\egroup }{2015}]{PGPD}
Jun Xu, Lei Zhang, Wangmeng Zuo, David Zhang, and Xiangchu Feng.
\newblock Patch group based nonlocal self-similarity prior learning for image
  denoising.
\newblock In {\em CVPR}, pages 244--252, 2015.

\bibitem[\protect\citeauthoryear{Xu \bgroup \em et al.\egroup
  }{2017}]{xu2017multi}
Jun Xu, Lei Zhang, David Zhang, and Xiangchu Feng.
\newblock Multi-channel weighted nuclear norm minimization for real color image
  denoising.
\newblock 2017.

\bibitem[\protect\citeauthoryear{Zhang \bgroup \em et al.\egroup
  }{2017a}]{zhang2017beyond}
Kai Zhang, Wangmeng Zuo, Yunjin Chen, Deyu Meng, and Lei Zhang.
\newblock Beyond a gaussian denoiser: Residual learning of deep cnn for image
  denoising.
\newblock {\em TIP}, 2017.

\bibitem[\protect\citeauthoryear{Zhang \bgroup \em et al.\egroup
  }{2017b}]{zhang2017learning}
Kai Zhang, Wangmeng Zuo, Shuhang Gu, and Lei Zhang.
\newblock Learning deep cnn denoiser prior for image restoration.
\newblock In {\em CVPR}, July 2017.

\bibitem[\protect\citeauthoryear{Zoran and Weiss}{2011}]{zoran2011learning}
Daniel Zoran and Yair Weiss.
\newblock From learning models of natural image patches to whole image
  restoration.
\newblock In {\em ICCV}, pages 479--486. IEEE, 2011.

\end{thebibliography}

\end{document}